%% file: paper.tex
\documentclass[letterpaper, 10 pt, conference]{ieeeconf}  %

\IEEEoverridecommandlockouts                              %

\input{tommiDefinition}

\usepackage{booktabs}

\title{\LARGE \bf
Avatarm: an Avatar With Manipulation Capabilities for the Physical Metaverse}

\author{A. Villani$^{1,*}$, G. Cortigiani$^{1,*}$, B. Brogi$^{1,*}$, N. D'Aurizio$^{1}$, T. Lisini Baldi$^{1,2}$, and D.~Prattichizzo$^{1,2}$%
\thanks{The research leading to these results has received funding from the European Union’s Horizon Europe programme under grant agreement No. 101070292 of the project \lq\lq HARIA - Human-Robot Sensorimotor Augmentation - Wearable Sensorimotor Interfaces and Supernumerary Robotic Limbs for Humans with Upper-limb Disabilities".}%
\thanks{$^1$ are with the Department of Information Engineering and Mathematics, University of Siena, Via Roma 56, 53100 Siena, Italy. \{villani, brogi, cortigiani, ndaurizio, lisini, prattichizzo\}@diism.unisi.it}%
\thanks{%
$^2$ are with the department of Humanoids and Human Centered Mechatronics (HHCM), Istituto Italiano di Tecnologia, Genova, Italy. %
}
\thanks{$^*$ These authors contributed equally to this work.}
}

\begin{document}

\maketitle 
\thispagestyle{empty}
\pagestyle{empty}

\begin{abstract}
Metaverse is an immersive shared space that remote users can access through virtual and augmented reality interfaces, enabling their avatars to interact with each other and the surrounding. Although digital objects can be manipulated, physical objects cannot be touched, grasped, or moved within the metaverse due to the lack of a suitable interface. 

This work proposes a solution to overcome this limitation by introducing the concept of a Physical Metaverse enabled by a new interface named “Avatarm”. The Avatarm consists in an avatar enhanced with a robotic arm that performs physical manipulation tasks while remaining entirely hidden in the metaverse. The users have the illusion that the avatar is directly manipulating objects without the mediation by a robot. The Avatarm is the first step towards a new metaverse, the “Physical Metaverse,” where users can physically interact each other and with the environment.

\end{abstract}

\section{Introduction}
\label{SEC:intro}

The word \lq\lq Metaverse" is a portmanteau of the words \lq\lq meta", which means beyond, and \lq\lq universe", which refers to everything that surrounds us, in the most general sense. In its simplest definition, the metaverse is a three-dimensional virtual world where social, cultural, and recreational activities take place as in the \textit{real} world. A more complex definition incorporates the use of Extended Reality (XR) technologies, \ie Augmented, Mixed and Virtual Reality, to either fully immerse or superimpose the user in a digital environment that enhances or replaces the physical reality of its body and its surroundings \cite{kozinets2022immersive}.

Started as a game-oriented application, today its popularity is rapidly growing thanks to a pool of users which is getting ever more familiar with the \lq digital lifestyle', laying fertile ground for its commercialization as a broader application for work and life \cite{forbesMeta}. Thanks to the metaverse, people may share the environment and interact with each other even if physically located in remote places, being engaged in a more immersive experience than the one offered by standard teleconferencing tools. Indeed, video-calls are able to stream audiovisual contents, but fail to convey the sense of presence as they lack the transmission of in-person interaction components, including body language and physical cooperation. In this regards, researchers are tackling the challenge of integrating the haptic channel together with the audio and video streams \cite{prattichizzo2010remotouch,baldi2017gesto, huisman2021feedback,hossain2011measurements,sanfilippo2015fully}, and their results offer a great potential for the metaverse too.

Adding the tactile layer is beneficial for increasing the realism of the interaction, as enabling users to manipulate objects in the shared environment shifts the participant's role from observer to actor. In addition, it has been demonstrated that collaborating with shared objects improves the quality and the efficacy of the communication  \cite{eckert2003role,brereton2000observational,nicolini2012understanding,bueger2017actor}.

While manipulating virtual objects is already possible thanks to outcomes of research in haptics, the physical interaction in the metaverse is not possible due to the lack of suitable interfaces. %

The following example will help to introduce the concept of \textit{Physical Metaverse} proposed in this paper. Suppose two friends are physically located in their respective homes, playing chess in the metaverse. They are both seated at a table in their living room, in front of their chessboards. They wear a head-mounted display (HMD) to see their environment in augmented reality, thanks to which they can see the opponent's avatar sitting in front of them. In the current implementation of the metaverse, what would happen in this situation is that both players would verbalize the desired move to the opponent, in order to allow each player to make both turns on their chessboard. But what if the avatar had manipulation capabilities and could actually \lq move the king' instead of asking the opponent to do it in its place? Being able to act on the physical objects of the shared environment broadens the horizon of the metaverse towards a new paradigm for physical interaction in XR.

\begin{figure}
\centering
\includegraphics[width=0.9\columnwidth]{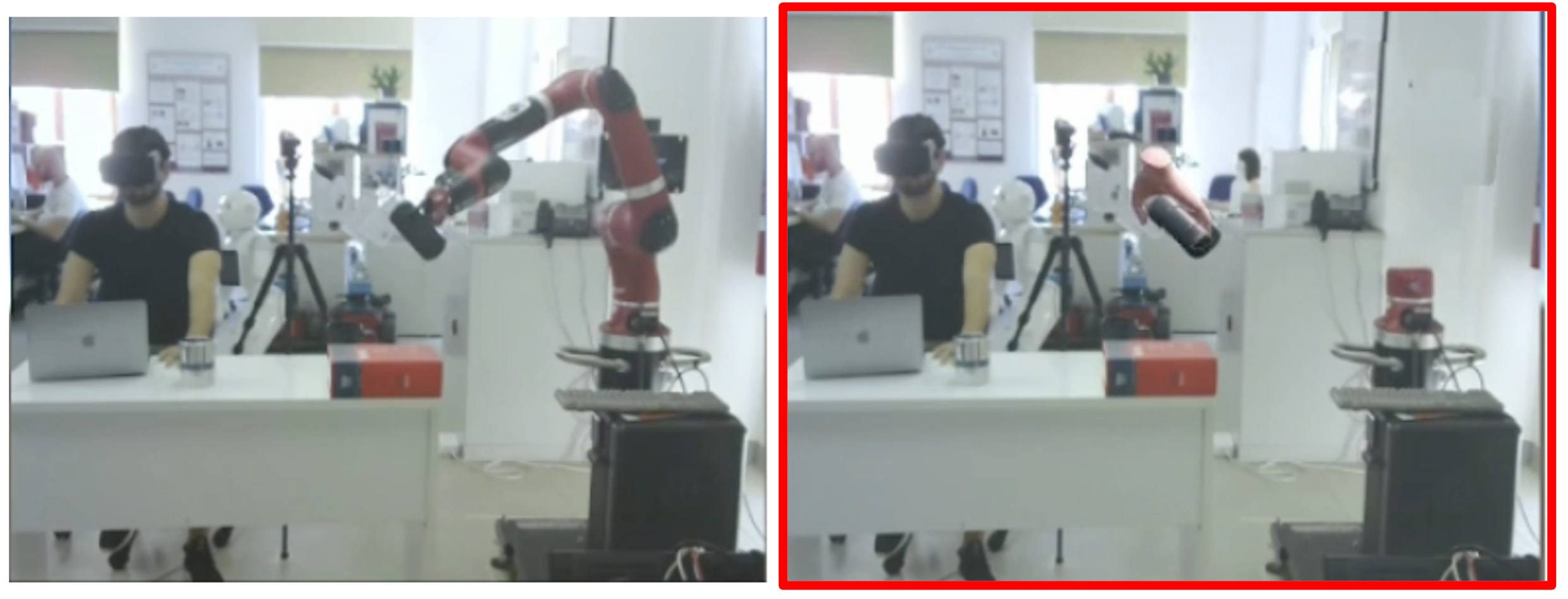}
\caption{Physical Metaverse representative scenario. On the left the scene captured by the real camera, on the right the scene as seen in the Physical Metaverse.}\label{Fig:Concept}
\end{figure}

In this paper, we lay the foundations for this ambitious scenario by proposing a novel interface, the \textit{Avatarm}: an avatar enhanced with a robotic arm that performs physical manipulation tasks while remaining entirely masked in the metaverse, as perceived by the users, to create the illusion that the user's avatar is directly manipulating objects without mediation by a robot (see \fig\ref{Fig:Concept}). Avatarms will populate a new generation of the metaverse, that we will define as {Physical Metaverse}. The real object manipulation is allowed by coupling the avatar actions with the actions of a robotic arm performing the physical manipulation task. The main idea of the Physical Metaverse is to hide the robot from view and replace it in the video stream with the hand of the avatar. The set of methodologies for
masking, removing, and seeing through objects in real time to diminish the reality is called Diminished Reality (DR). The interested reader may find a comprehensive review in \cite{mori2017survey}. However, very few works apply DR in human-robot collaboration scenarios. They are mostly thought to allow users to see occluded areas behind the robot \cite{taylor2020diminished}, without the metaverse application in mind. In addition, differently from the existing vision-based solutions, in this work we propose a technique to determine the region of interest by using the CAD model and the kinematics of the robot.

In what follows, we present the steps leading to the implementation of the Avatarm, as well as a demonstrative scenario in the Physical Metaverse.

\section{Physical Metaverse\label{SEC:Meta}}
The metaverse is a multi-user environment merging physical reality with digital virtuality. 
In its current form, the metaverse enables real-time user communication and interaction with digital entities. The key feature that is currently missing is the physical interaction with real remote objects.

\begin{figure}
\centering
\includegraphics[width=0.85\columnwidth]{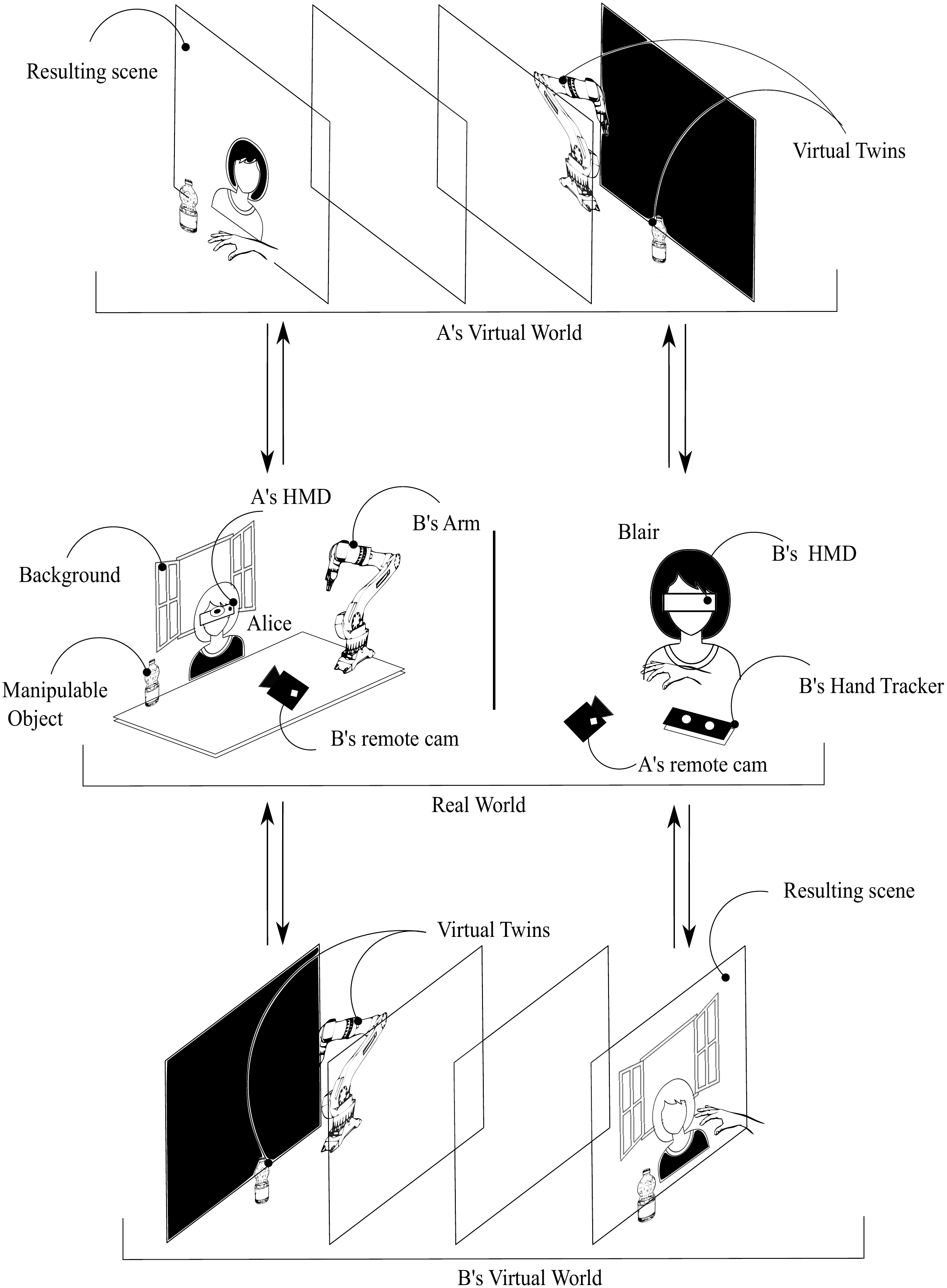}
\caption{An illustrative scenario of the Physical Metaverse. Two users, Alice (A) and Blair (B), are physically remote and can interact in the metaverse, sharing an augmented version of A's physical environment. Alice observes her workspace through a head-mounted display, while B observes the same environment from a remote camera that streams into her own HMD. B is the Avatarm of the scenario, hence she can manipulate real objects in A's physical environment exploiting the robotic arm placed on A's desk. In the virtual world, the robot is removed from the video stream and substituted with B's virtual hand thanks to an overlap of panels.
}
\label{Fig:Idea}
\end{figure}

To overcome this limitation and empower the user with touch ability, we developed a framework to enhance the virtual avatar with manipulation capabilities, transforming the avatar into the Avatarm. For the sake of simplicity, the following scenario (represented in \fig\ref{Fig:Idea})  will be instrumental from now on in explaining and characterizing the framework of the Physical Metaverse.

 Let us consider two people, Alice and Blair, who meet in the metaverse in a shared environment. %
In particular, Alice is seated at her desk wearing an HMD, and her room is equipped with a camera and a robotic arm that can manipulate the objects on the table. The robotic arm is remotely controlled by Blair, which will be the Avatarm in the scenario: the robotic arm represents the \lq\lq Blair-arm". 
The shared environment is  Alice's physical environment augmented with the avatar of Alice (as she does not need to manipulate remote objects) and the Avatarm of Blair, and diminished of the robotic arm. Being a shared environment, the scene is rendered in both HMDs, but each participant observes the scene from its own point of view: Alice uses the camera mounted in her HMD, while Blair uses the additional camera in Alice's room, which is placed in front of the desk. When Blair manipulates an object on Alice's desk, the interaction between the virtual hand of Blair and the real object is handled through a digital twin of the latter. Hence, Alice and Blair see the virtual hand of Blair grasping and moving the object digital twin, while in the meantime the robotic arm is moving the real object. 

To let the Physical Metaverse work, Blair's hand tracking is needed to control both the robotic arm and the virtual hand, while the robotic arm transmits its pose to enable the real-time cancellation of its structure. 

In the rest of the section, we describe in detail the building blocks for the Physical Metaverse.

\begin{figure}
\centering
\includegraphics[width=1\columnwidth]{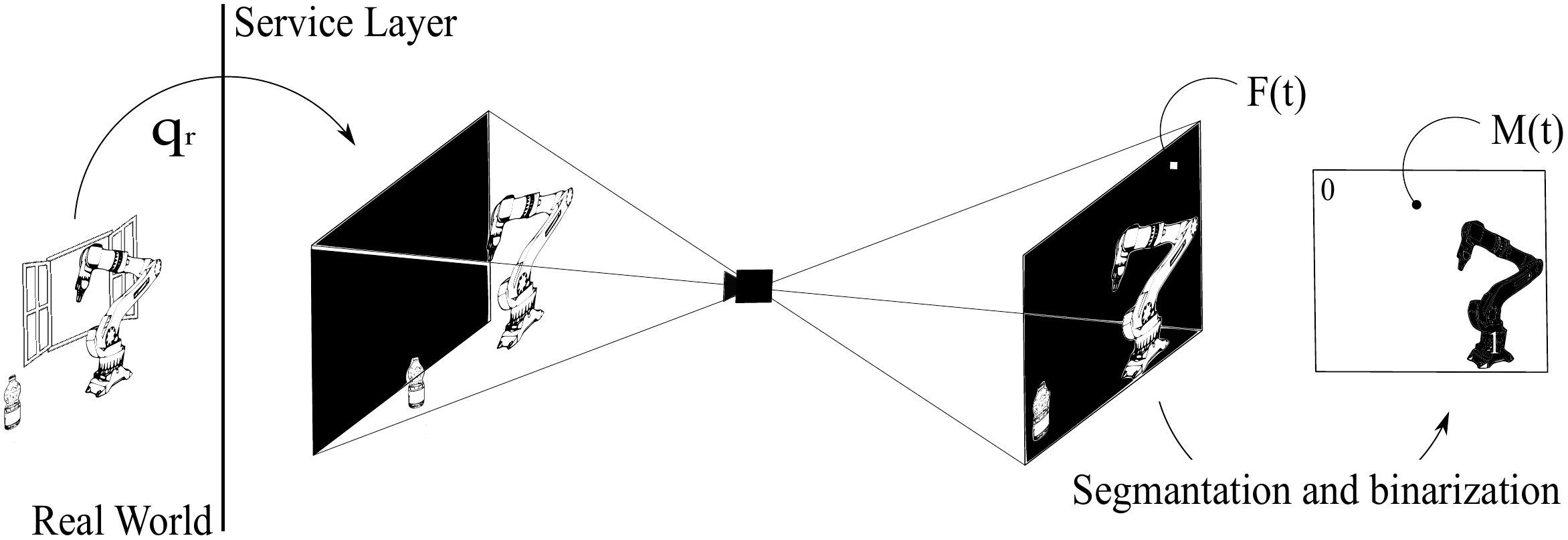}
\caption{Service layer functioning. The robotic arm transmits its pose to the monochromatic digital twin, which in turn updates its pose. A virtual camera, having the same
optical and spatial features of the real camera, records the digital twin while it is moving in front of a black panel. The segmentation and binarization of the image frames generate the mask to hide the robot in the interaction layer.}\label{Fig:SLayer}
\end{figure}

\subsection{Robot Cancelling Method}\label{SEC:VR}
In the proposed framework, the real-time cancellation of the robotic arm is performed through a developed 3D world. The software stack is composed of two layers, namely the \textit{Service Layer} and the \textit{Interaction Layer}.

\pg{Service Layer}
The service layer is the lowest layer of this architecture and is in charge of %
creating a mask with the region of the scene covered by the robot. %
This is done by combining video processing techniques with data coming from the robot kinematics. The layer functioning is visually depicted in \fig\ref{Fig:SLayer}. 

As first step, we generated a digital twin of the robotic arm using its CAD model. This allowed to mimic the robot kinematics in the virtual world by using the real joint values measured during the task execution. Then, to identify the portion of the scene covered by the robotic arm, we first set the color of the digital twin as white, then we placed the latter in front of a black panel, with a virtual eye recording the scene from a point of view chosen to resemble the field of view of the real camera with respect to the real robot. The focal length ($\varphi$) of the virtual camera ($v$) in the service layer ($s$) is set equal to the one of the real camera ($r$), \ie $\varphi_{v,s}=\varphi_{r}$. Two video sources are hence involved in the cancellation process:
the virtual eye that acquires frames of the robotic arm digital twin, and the real camera that streams real world images. 
The image frames of the videos acquired by both cameras are expressed as time-varying 3D matrices. For the virtual camera, the matrix is denoted with $F(t) \in \Re^{X\times Y\times C}$, where $X\times Y$ corresponds to the camera resolution, and $C$ is the cardinality of the colour space. %
In this work, image frames are encoded using RGB$\alpha$ values ($C=4$), %
thus a single pixel $\wp$ of $F(t)$ can be expressed as:
\begin{equation*}
\wp_F(x,y,t)=[r_F(x,y,t),g_F(x,y,t),b_F(x,y,t), \alpha_F(x,y,t)] 
\end{equation*}
being $(x,y)$ $ \forall x,y\leq X,Y$ the pixel coordinates, $r$, $g$, and $b$ the three colour channels, and $\alpha$ the channel for the matte information. 
The segmentation and binarization algorithm creates a binary mask $M(t) \in \Re^{X\times Y}$ whose pixels $\wp_M$
assume at time $t$ the value $m_{x,y}(t)=1$ if $\wp_F(x,y,t)$ contains a portion of the robot image, the value $m_{x,y}(t)=0$ otherwise. The uniformity predicate adopted for the region-based segmentation algorithm requires that a pixel $\wp_F$ contains a portion of the robot if its RGB$\alpha$ values are in the ranges $\tilde{r}=[r_{min}, r_{max}]$, $\tilde{g}=[g_{min}, g_{max}]$, $\tilde{b}=[b_{min}, b_{max}]$, and $\tilde{\alpha}=[\alpha_{min}, \alpha_{max}]$, respectively. Thus, 
\begin{equation*}
\wp_M(x,y,t)= \begin{cases} 1 & \mbox{ if } \wp_F(x,y,t)  \in \tilde{r}\cap\tilde{g}\cap\tilde{b}\cap\tilde{\alpha} \\ 0 & \mbox{ otherwise.}\end{cases}
\end{equation*}

\begin{figure}[t]
\centering
\includegraphics[width=0.85\columnwidth]{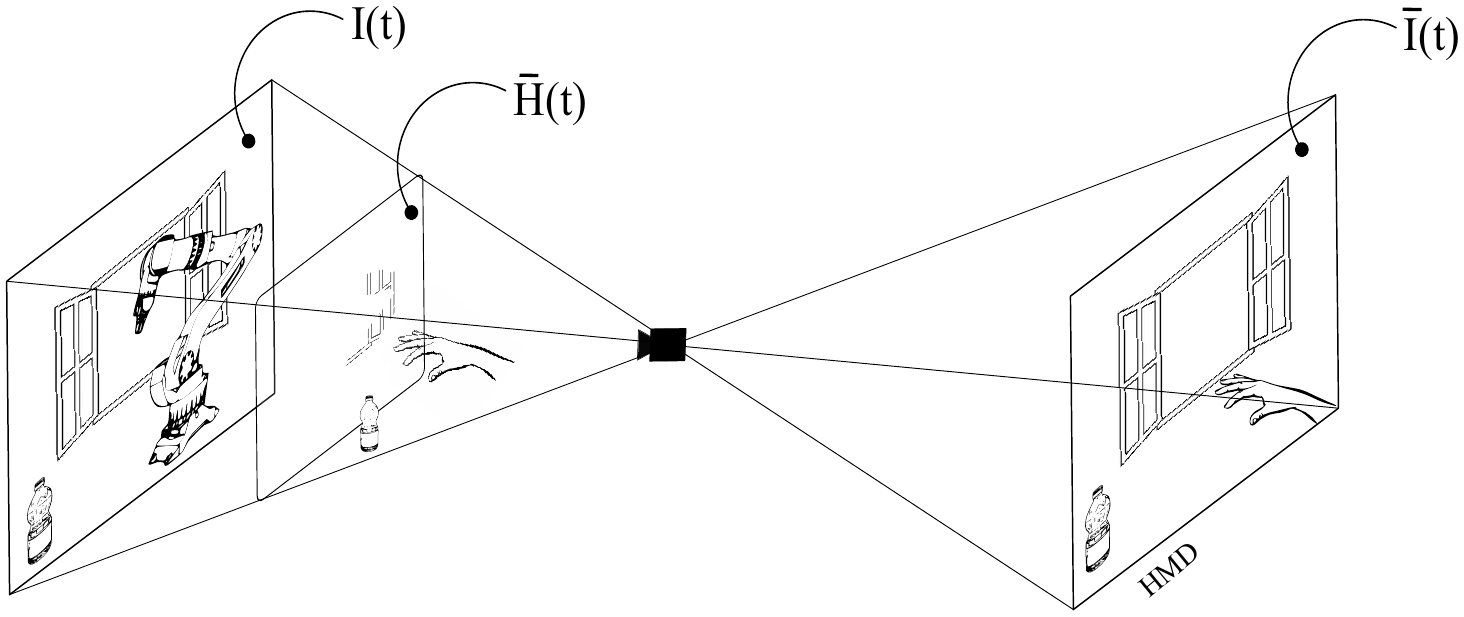}
\caption{Interaction layer functioning. Image frames of the real camera are streamed on a virtual panel, which is overlapped by a semi-transparent panel showing the background image filtered with the mask computed in the service layer. The real object digital twin and the virtual hand are conveniently scaled and located between the panels and the virtual camera.}\label{Fig:ILayer}
\end{figure}
\pg{Interaction Layer}
The interaction layer (see \fig\ref{Fig:ILayer}) uses the output of the service layer, \ie the mask, and a background image (acquired before placing the robotic arm in the real environment) to make the robot transparent in the shared environment. 

To this end, the image frames acquired by the real camera $I(t) \in \Re^{X\times Y\times C}$ are projected on a virtual panel. The latter is overlapped by a second virtual panel, which is used to stream a new 3D matrix $\tilde{H}\in\Re^{X\times Y\times C}$ that is a combination of the background image $H\in\Re^{X\times Y\times C}$ and the time-dependent binary mask $M(t)$. Thus, pixels $\wp_{\tilde{H}}$ are computed as:
\begin{equation*}
\wp_{\tilde{H}}(x,y,t)=\left[ r_H(x,y),  g_H(x,y),  b_H(x,y), \alpha_H(x,y)m_{x,y}(t) \right].
\end{equation*}
Thanks to the panels overlapping, the two images $I(t)$ and $\tilde{H}(t)$ are fused together, generating the image $\tilde{I}(t)$, whose pixels result:
\begin{equation*}
\wp_{\tilde{I}}(x,y,t)=\begin{cases}\wp_{I}(x,y,t) & \forall x,y : m_{x,y}(t)=0; \\ \wp_{H}(x,y)  & \forall x,y : m_{x,y}(t)=1.\end{cases}
\end{equation*}
At this point, the robotic arm is made transparent in $\tilde{I}(t)$ by replacing the pixels within the mask with portions of the background image. The resulting manipulator-free image is firstly recorded by the virtual camera simulating the user point of view, then superimposed by the Avatarm's virtual hand and by the digital twin of the real object to be manipulated, and finally rendered into the HMD.

As regards the object digital twin, enabling its manipulation in the virtual environment is not a straightforward operation. Indeed, the virtual object should have not only the same aspect but also the same relative position with respect to the camera and the same size as the real object. Since in our framework the scene is projected on a virtual panel, having the relative positions of the two objects coincident would mean that the virtual object disappears behind the panel when the user moves the object away from the camera. To solve this problem, we changed the relative position of the virtual object and we let the dimensions of the two objects coincide when the virtual object is in front of the virtual panel, far enough to avoid it crossing the panel during the interaction. Hence, the position of the digital twin center $o_v=[x_v, y_v ,z_v]^T$ and the scaling factor to compute its dimension $s_v$ are evaluated according to the perspective transformation for overlapping the pixels $\wp_I(x',y',t)$ that contain portions of the real object. In particular, once the depth $z_v$ has been defined (according to the manipulation workspace), the 2D projection of the virtual object center on the vertical plane, \ie $x_v'$ and $y_v'$, and the respective scaling factor $s_v'$ need to satisfy the following equations:
\begin{equation*}\label{eq:collocation}
\left[\begin{matrix}x'\\ y'\end{matrix}\right]-\left[\begin{matrix} \frac{(x_v'+1)X}{2s_v'}\\ \frac{(y_v'+1)Y}{2s_v'}\end{matrix}\right]=0
\end{equation*}
 \begin{equation*}
\left[\begin{matrix}x_v'\\ y_v'\\z_v'\\s_v'\end{matrix}\right] = \left[\begin{matrix}\frac{Y}{X}\varphi_{v,i}  &0 &0 &0\\ 0 &\varphi_{v,i} &0 &0 \\ 0 & 0 & \frac{far+near}{near-far} &2\frac{far \cdot near}{near-far}\\0 & 0 & -1&0 \end{matrix}\right]\left[\begin{matrix}x_v\\ y_v\\z_v\\s_v\end{matrix}\right]
\end{equation*}
where $\varphi_{v,i}$ is the focal length of the virtual camera in the interaction layer, $far$ and $near$ are the distances between the camera and the nearest plane, and the camera and the farthest plane of the viewing frustum, respectively. The 2D projection holds under the hypotheses of camera reference systems alignment.

\subsection{Robot Motion Control}
\label{SEC:Robot}
In our framework for the Physical Metaverse, the error between the positions of the real object and its digital twin during the interaction can be minimized if the robotic arm has the ability to anticipate the human action by understanding which is the target object and preventively grasp it. For the sake of practicality and since this feature is not relevant at this stage of the work, we decided not to implement it. Indeed, this capability is not a novelty in robotics, several studies already demonstrated its feasibility \cite{miller2003automatic, lin2015robot, kappler2015leveraging,saponaro2013robot,miao2015optimal,gasparetto2015path}. Differently, implementing the algorithm %
to map the motion of the real human hand into the Avatarm's hand is of interest to demonstrate the potential value of the Physical Metaverse.

\begin{figure}
\centering
\includegraphics[width=0.9\columnwidth]{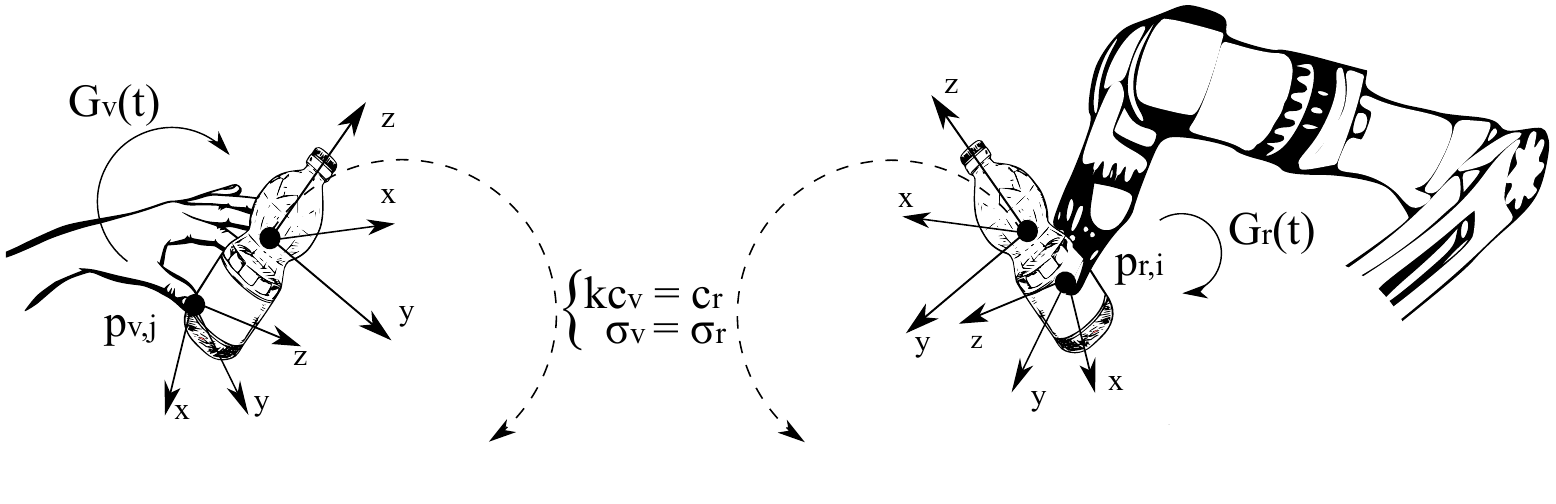}
\caption{Starting from the virtual contact points, the dynamics at the gravity center of the digital twin is evaluated through the grasping matrix and mapped into the physical counterpart. These values will be the reference of the robotic arm. }\label{Fig:Mapping}
\vspace{-2 ex}
\end{figure}

To ensure the realism of the interaction, we sought to match movements and forces applied on the gravity centers of the real object and its digital twin. Thus, we implemented a mapping algorithm based on the contact effects on the two objects. When the virtual hand grasps the digital twin, from the virtual scene we can compute at each time instant $t$ both position $p_{v,j}(t)$ and orientation $\theta_{v,j}(t)$ of each contact point $j$, as well as forces $f_{v,j}$ and torques $\tau_{v,j}$ applied to the virtual object. According to the grasping theory \cite{prattichizzo2016grasping}, the linear relationship between the forces and torques applied at the contact points, and the external forces $\hat{f}_{v}$ and torques $\hat{\tau}_{v}$ applied to the gravity center of the object is given by the grasp matrix $G_v(t)$: 
\begin{equation*}
\lambda_v(t) =    \left[\begin{matrix}f_{v,1}\\ \tau_{v,1} \\ \vdots  \\ f_{v,n} \\ \tau_{v,n}\\ \end{matrix}\right] =   \underbrace{\left[\begin{matrix}\overline{R}_{v,1} P_{v,1}\\ \vdots \\ \overline{R}_{v,n} P_{v,n}\\ \end{matrix}\right]}_{G_v(t)} w_v(t),
\end{equation*}
where $\lambda_v$ is the vector of the generalized forces applied at the $n$ contact points, and $w_v=\left[\begin{matrix}\hat{f}_{v}& \hat{\tau}_{v}  \end{matrix}\right]^T$ is the wrench vector. The matrix $P_{v,j}$ describes the effects (at the gravity center) of the variations of position and orientation of the $j$-th contact point, expressed in contact point reference frame. $\overline{R}_{v,j}$ is a block diagonal matrix with the orientation of the $j$-th contact frame with respect to the inertial frame ${R}_{v,j}$, \ied
 \begin{equation*}
P_{v,j}=\left[\begin{matrix}I_{3\times3} & 0_{3\times3}  \\ S(p_{v,j}(t)-\hat{p}_v(t))  & I_{3\times3}\\ \end{matrix}\right]
\overline{R}_{v,j}=\left[\begin{matrix}{R}_{v,j} & 0_{3\times3}  \\ 0_{3\times3}  & {R}_{v,j}\\ \end{matrix}\right],
\end{equation*}
where $S(p_{v,j}(t)-\hat{p}_v(t))$ is the cross product matrix of the position of the contact point and the position of the gravity center $\hat{p}_{v}$. The squeezing contribution of each contact point $\xi_v$  on the virtual object is evaluated by computing the internal forces, \ie forces that do not contribute to the acceleration of the object:
\begin{equation}\label{eq:Null}
\xi_v(t) =\underbrace{\left( I_{6n\times6n}-G_v^TG_v\right)^\dagger}_{N(G_v)^\dagger}\lambda_v(t),
\end{equation}
being $N(G_v)$ the projection into the null space of $G_v(t)$\footnote{$(\cdot)^\dagger$ denotes the pseudoinverse operation.}. The resulting squeezing force $\sigma_v$ on the object is computed as the mean value of the squeezing contribution of each contact point:
\begin{equation}\label{eq:Mean}
\sigma_v(t) =\underbrace{\frac{1}{n}\left[ \begin{matrix} I_{6\times6}& \dots & I_{6\times6}\end{matrix}\right]_{6\times 6n}}_{\Gamma_v} \xi_v(t). 
\end{equation} 
By means of the kinetostatic duality, the generalized velocity at the object center of gravity $\dot{c}_v(t)=\left[\begin{matrix}\dot{\hat{p}}_v(t)&\hat{\omega}_v(t) \end{matrix}\right]^T$ is computed as:
\begin{equation}\label{eq:Dual}
\dot{c}_v(t) =\left[\begin{matrix}\overline{R}_{v,1}^T P_{v,1}^T & \dots & \overline{R}_{v,n}^T P_{v,n}^T\\ \end{matrix}\right]\nu_v(t) = G_v(t)^T \nu_v(t)
\end{equation}
where $\nu_v(t) = \left[\dot{p}_{v,1}, \omega_{v,1}, ..., \dot{p}_{v,n}, \omega_{v,n}\\ \right]^T$ is the generalized velocity vector of the $n$ contact points. As a result, we can express the relation between the generalized forces and velocities at the contact points and at the gravity center as:
\begin{equation}\label{eq:Virtual}
\left[\begin{matrix} \sigma_v(t) \\ \dot{c}_v(t)\end{matrix}\right]=\left[\begin{matrix}\Gamma_v N(G_v)^\dagger & 0  \\ 0 & G_v^T\end{matrix}\right]\left[\begin{matrix} \lambda_v(t) \\ \nu_v(t)\end{matrix}\right].
\end{equation}

Once the movements and forces of the digital twin are known, 
these are mapped one-to-one to the real object, which is grasped by the robotic arm (see \fig\ref{Fig:Mapping}). By denoting with $\sigma_r(t)$ and $\dot{c}_r(t)$ the squeezing force and the generalized velocity at the gravity center of the real object, we set:
\begin{equation}\label{eq:Rel}
\left[\begin{matrix} \sigma_r(t) \\ \dot{c}_r(t)\end{matrix}\right]=\left[\begin{matrix} 1 &0 \\ 0&k\end{matrix}\right] \left[\begin{matrix} \sigma_v(t) \\ \dot{c}_v(t)\end{matrix}\right]
\end{equation}
where $k$ is a scaling factor needed to compensate the motion disparity due to the focal length difference between the virtual camera $\varphi_{v,i}$ and real one $\varphi_r$, \ied
\begin{equation*}
k=\left[ \begin{matrix} \frac{\varphi_r}{\varphi_{v,i}} I_{2\times2} & 0_{2\times1} & 0_{2\times3} \\ 0_{1\times2} & 1 & 0_{1\times3} \\  0_{3\times3} & 0_{3\times1}  &  I_{3\times3}    \end{matrix}\right].
\end{equation*}
According to \eqref{eq:Null}, \eqref{eq:Mean}, and \eqref{eq:Dual}, the vectors of the generalized forces $\lambda_r$ and velocities $v_r$ applied by the $m$ contact points to the real object are given as:
\begin{equation}\label{eq:Real}
\left[\begin{matrix} \lambda_r(t) \\ \nu_r(t)\end{matrix}\right]= \left[\begin{matrix} N(G_r)\tilde{\Gamma}_r & 0  \\ 0 &( G_r^T)^\dagger\end{matrix}\right] \left[\begin{matrix} \sigma_r(t) \\ \dot{c}_r(t)\end{matrix}\right] 
\end{equation}
where $G_r \in\Re^{6m\times6}$ and $\tilde{\Gamma}_r \in\Re^{6m\times6}$ are defined as:
\begin{equation*}
G_r=\left[ \begin{matrix} \begin{matrix}R_{r,1} & 0  \\ {R}_{r,1} S(p_{r,1}-\hat{p}_r) &  R_{r,1}\\ \end{matrix} \\ \vdots \\ \begin{matrix}{R}_{r,m} & 0  \\ {R}_{r,m} S(p_{r,m}-\hat{p}_r)    & {R}_{r,m}\\ \end{matrix}\end{matrix} \right]
\tilde{\Gamma}_r=m \left[ \begin{matrix} I_{6\times6}\\ \vdots \\ I_{6\times6}\end{matrix}\right]_{6m\times6}.
\end{equation*}
By substituting \eqref{eq:Virtual} and \eqref{eq:Rel} in \eqref{eq:Real}, we can express $\lambda_r$ and $v_r$ as a function of $\lambda_v$ and $v_v$, \ied
\begin{equation*}
\left[\begin{matrix} \lambda_r(t) \\ \nu_r(t)\end{matrix}\right]= \left[\begin{matrix} N(G_r)\tilde{\Gamma}_r\Gamma_vN(G_v)^\dagger & 0  \\ 0 & (G_r^T)^\dagger k G_v^T \end{matrix}\right] \left[\begin{matrix} \lambda_v(t) \\ \nu_v(t)\end{matrix}\right].
\end{equation*} 
Recalling the kinetostatic duality, we compute the joint reference velocities $\dot{q}_r$ and torques $\tau_{q,r}$ as:
\begin{equation*}
\left[\begin{matrix} \tau_{q,r}(t) \\ \dot{q}_r(t)\end{matrix}\right]= \left[\begin{matrix} J(q_r)^T& 0  \\ 0 & J(q_r)^\dagger  \end{matrix}\right]  \left[\begin{matrix} \lambda_r(t) \\ \nu_r(t)\end{matrix}\right]
\end{equation*} 
where  $J(q_r)$ is the Jacobian matrix of the robotic arm.

\begin{figure}
\centering
\includegraphics[width=0.80\columnwidth]{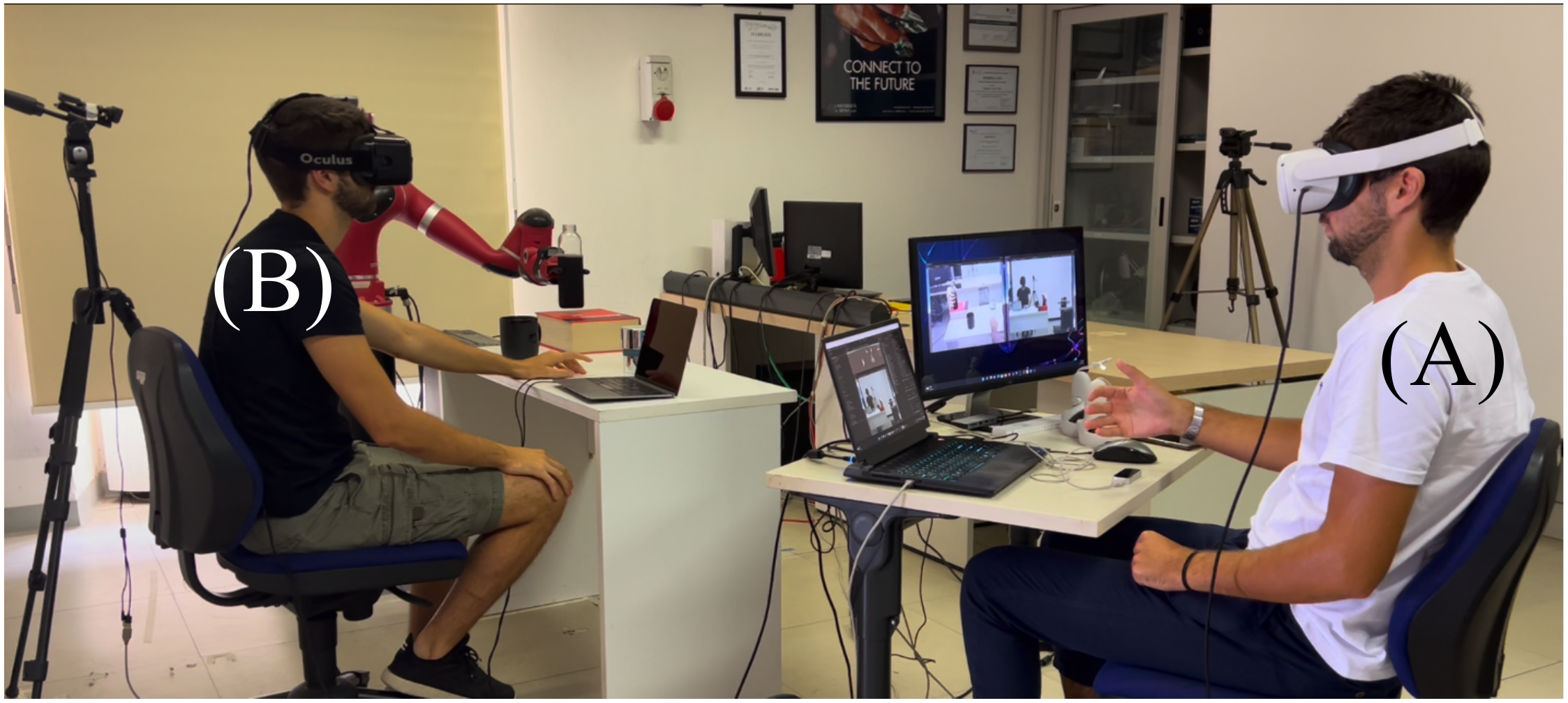}
\caption{Laboratory setting for the practical implementation of the Physical Metaverse. The user A pours the water in B's cup thanks to the Avatarm, while B observes the scene. }\label{Fig:RobotSetUp}
\end{figure}

\section{Practical implementation} \label{SEC:Implementation}

To demonstrate the feasibility of the proposed approach, we tested the system in the scenario of Alice and Blair presented in \sect\ref{SEC:Meta}. In particular, as representative task we choose Blair using her Avatarm to pour a glass of water to Alice. The living lab setting is reported in \fig\ref{Fig:RobotSetUp}.

\begin{figure}
\centering
\includegraphics[width=0.8\columnwidth]{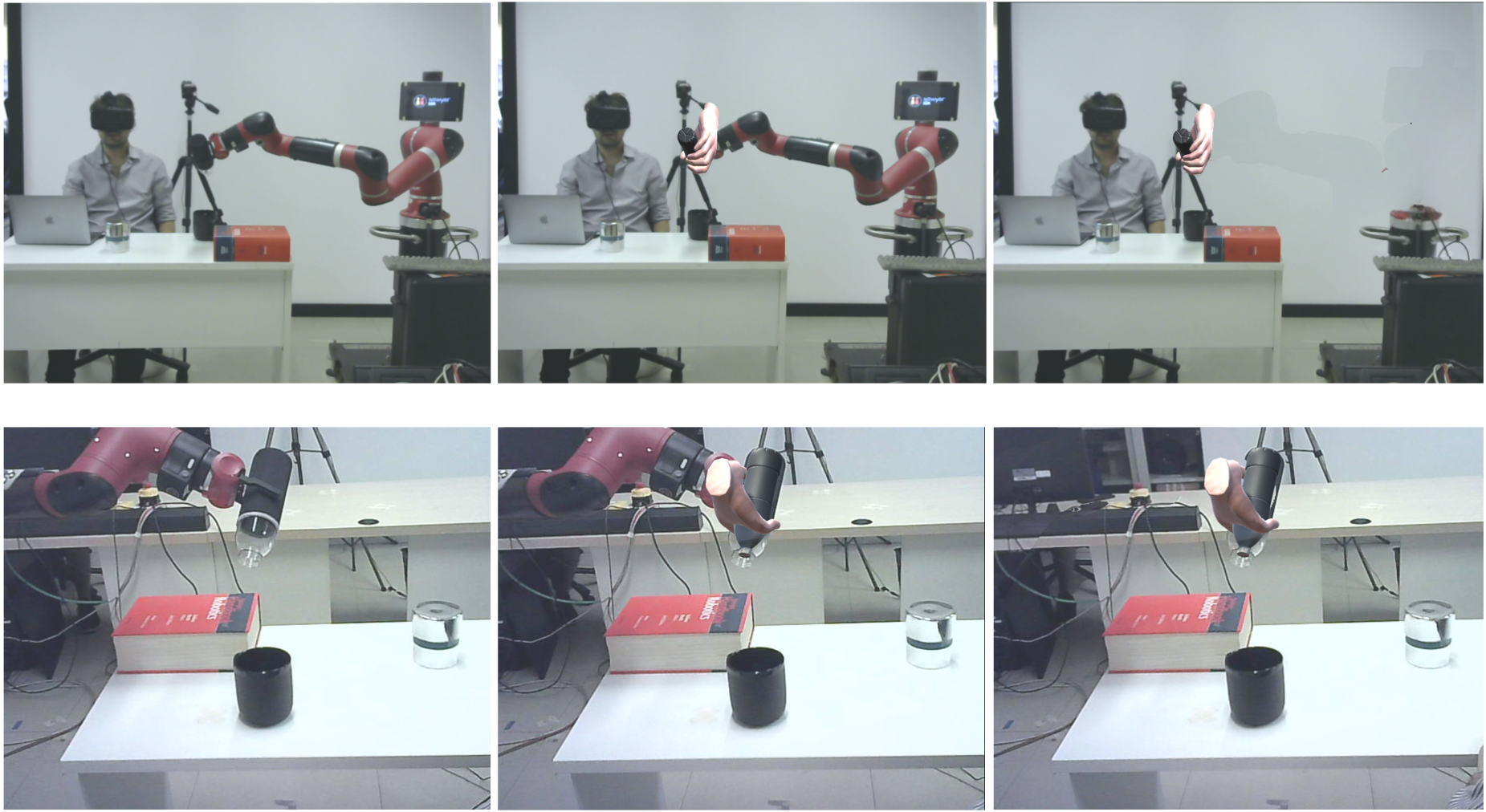}\\
\caption{Pouring task. %
The Physical Metaverse as seen from two users (top for user A, bottom for user B).  From left to right, the same image frame acquired by real cameras for both users, the image of the scene augmented with the virtual hand and the digital twin, and the image of the scene diminished of the robotic arm.
}\label{Fig:hiding}
\end{figure}

The experimental setup consists of a desk with some objects of common use, including an empty glass and a water bottle. The Avatarm is realized using a Sawyer (Rethink Robotics GmbH, DE). A maximum velocity constraint at \SI{0.25}{\m\per\s} was imposed on the robot movements to be compliant with the human-robot interaction standard (ISO-10218)  \cite{international2006robots}. Regarding the video sources, we exploited two full HD 1080P USB webcams. The shared environment and the two software layers, instrumental for the scene reconstruction, are implemented using Unity Graphic Engine (Unity Technologies, US), and displayed through Oculus Quest 2 (Meta Platforms, Inc., US). A Leap Motion Controller (UltraLeap inc., US) is used  to track the hand, whereas CLAP [23] is used for reproducing the interaction between the hand and the object in the virtual environment as described in \sect\ref{SEC:Robot}. ROS Melodic on Ubuntu 18.04 is used for connecting all the system components. For what concerns the digital environment, the mask $M(t)$ is computed using the frames acquired by an 8-bit RGB$\alpha$ virtual eye observing the CAD model of the robot.
The uniformity thresholds on the RGB$\alpha$ space are setted in accordance with the chromatic choices of the service layer and the colour depth of the virtual point of view. These values are reported in \tabb\ref{Tab:InfoVR} together with other relevant parameters. The results of the robot-diminishing are reported in \fig\ref{Fig:hiding}, and a video showing the working system is publicly available\footnote{ \url{http://sirslab.dii.unisi.it/videos/2022/ICRA2023.mp4}}.

\begin{figure}
\centering
\subfloat[]{
\includegraphics[width=0.8\columnwidth]{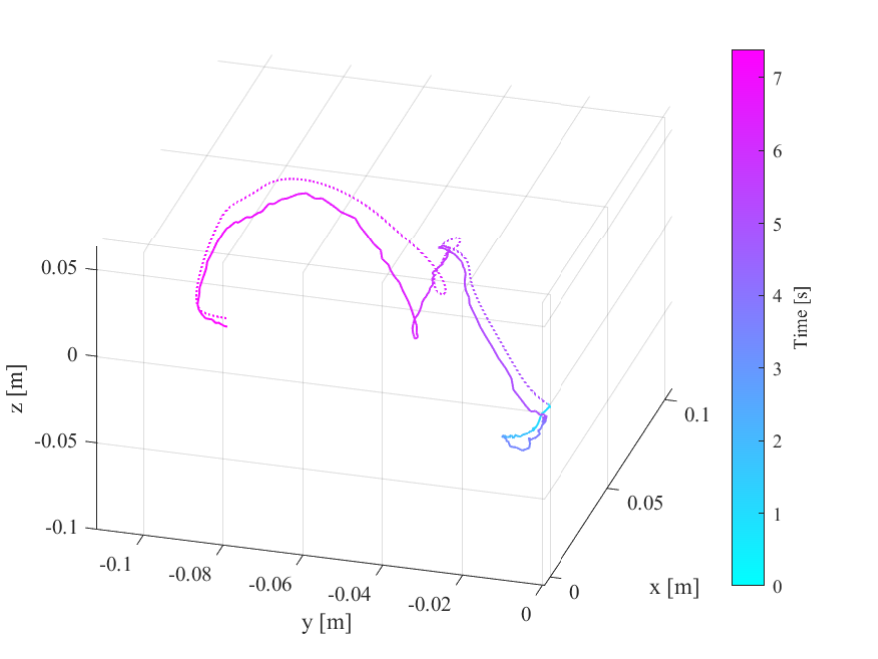}
\label{Fig:traj}}\\
\subfloat[]{
\includegraphics[width=0.9\columnwidth]{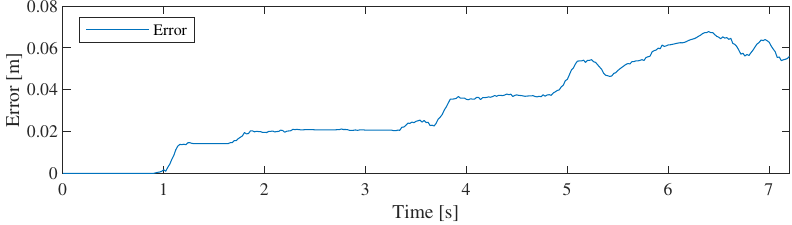}
\label{Fig:trajError}}\\
\subfloat[]{
\includegraphics[width=0.9\columnwidth]{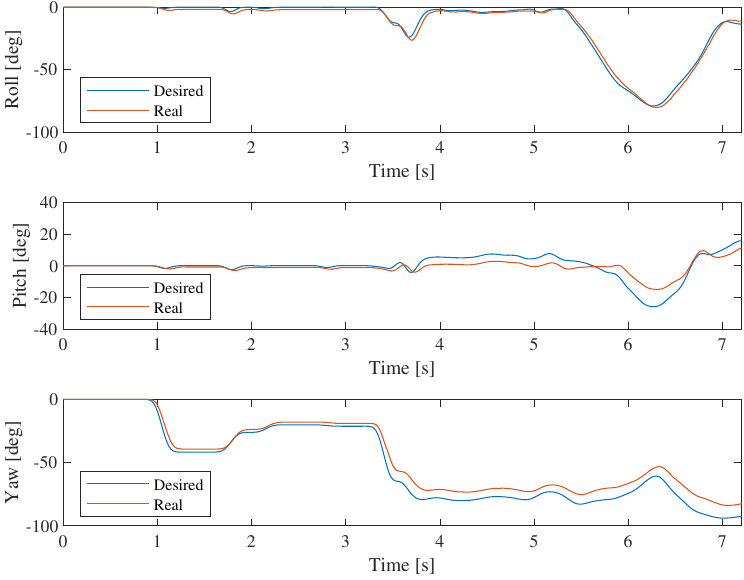}
\label{Fig:orient}}
\caption{Representative trial of the pouring task. The user grasps the bottle (initial part of the trial), lifts it at about 5 \si{\centi\m} from the desk, pours the water, and then moves the bottle 9 \si{\centi\m} above. In a), the trajectories of the real bottle of water and its digital twin. The path of the virtual object is depicted with a dotted line, while the solid line depicts the path followed by the real object. In b), the error in following the desired trajectory, i.e. the euclidean distance between the real and the virtual object centres of mass. In c), the orientation error computed as the root mean square error of roll (upper panel), pitch (middle panel), and yaw (lower panel). It is worth noticing that the error is computed between the real object and the digital twin, possible mismatches between the real hand and the virtual one due to the tracking system are not considered.}
\end{figure}

The pouring task was repeated 10 times, and the position and orientation errors between the real bottle of water and its digital twin were used as evaluation metrics.  All users completed the task successfully. In \fig\ref{Fig:traj} we report the paths obtained in a representative trial, while the corresponding error is reported in \fig\ref{Fig:trajError}. The resulting position error, computed as the root mean square error of the norm among all the trials, is $0.0173 \pm 0.0132$ \si{\m}.
Similarly, we computed the average orientation error among all trials as the root mean square error of roll, pitch and yaw, obtaining $1.81\pm 1.80$ \si{\deg}, $2.08\pm 2.49$ \si{\deg}, $3.99\pm 5.27$ \si{\deg}, respectively. Results of a representative trial are in \fig\ref{Fig:orient}.

\begin{table}
\scriptsize
  \centering
   \begin{tabular}{lcc}\hline
   \textbf{Parameter}  & \textbf{Description}  & \textbf{Value}  \\ \hline\hline
    $\tilde{r}=\tilde{g}=\tilde{b}=\tilde{\alpha}$ & \multicolumn{1}{c}{Range of RGB$\alpha$ levels of $F(t)$} & [240,255] \\
    $X\times Y$ & Sizes of images $F(t)$ and $\tilde{\Im}(t)$ in pixels & \multicolumn{1}{c}{640$\times$480} \\
    $\#C_v = \#C_r$ & Color space depth & \multicolumn{1}{c}{8 bit} \\
    $far$ &   Camera-nearest visible plane distance  & \SI{95}{cm} \\
    $near$ &   Camera-farest visible plane distance  & \SI{75}{cm} \\
    $\varphi_{v,i}$ & Focal length  & \SI{70}{cm} \\
     $\varphi_{v,s} = \varphi_r$ & Focal length  & \SI{75}{cm} \\\hline
    \end{tabular}%
      \caption{Parameters values used for the practical implementation of the Physical Metaverse.}
  \label{Tab:InfoVR}%
\end{table}%

\section{Conclusions}
With this work we introduced the Avatarms, a new generation of avatars with the ability to manipulate physical objects in the metaverse. Their implementation broadens the perspective of the digital interaction toward the Physical Metaverse, a 3D extended world where users have real object-mediated communication. This is made possible thanks to a robotic arm that manipulates the objects, and a software that hides the robot from the video and replaces it with the virtual hand.

Together with the definition of the scenario, we developed a working framework for the Physical Metaverse, providing an approach to diminish the shared environment of the robotic arm, enable the interaction with digital twins of real objects, and control the robot for manipulating the real object along the trajectory traced by the digital twin. An experiment in living lab settings has been conducted to demonstrate the feasibility of the Avatarm.

Some limitations of the current framework for the Physical Metaverse need to be acknowledged. The first one is the lack of algorithms for predicting the human manipulation intention, which could reduce the delay between the real and the virtual systems, hence improving the overall experience. Moreover, we have not added any haptic cue to the interaction, but we have reason to believe that the tactile layer will be easily integrated in the future using state-of-the-art solutions. Finally, implementing automatic contrast and brightness adjustment of the background image would improve the cancellation of the robot, making it robust to light variations in the room.

The results from the early stages of development of the Avatarm are promising, paving the way for further research into the implications of its use. Specifically, future work could explore how the Avatarm affects the feeling of presence and collaboration between two or more users engaging in interactive or assistive tasks. With this research, we may one day make the Physical Metaverse a reality.

\bibliographystyle{IEEEtran}
\bibliography{biblio/IEEEabrv,biblio/conference,biblio/biblio}

\end{document}

%% file: tommiDefinition.tex
\usepackage[usenames,dvipsnames]{xcolor}

\usepackage{epsfig} %
\usepackage{graphics} %
\usepackage{amsmath} %
\usepackage{amssymb}  %
\usepackage{cite}	
\usepackage{graphicx}
\usepackage{tabularx}
\usepackage{lipsum}
\usepackage{multirow}
\usepackage{psfrag}
\usepackage{float}
\usepackage{placeins}

\usepackage{algorithm}
\usepackage{algorithmicx}
\usepackage[]{algpseudocode}

\usepackage{subfloat}
\usepackage{subfig}
\usepackage{caption}

\usepackage{soul}

\usepackage[colorlinks = true,
            linkcolor = black,
            urlcolor  = black,
            citecolor = black,
            anchorcolor = black]{hyperref}

\usepackage{bm}
\usepackage{microtype}
\usepackage{wasysym} %
\usepackage{adjustbox}
\usepackage{lipsum}

\let\labelindent\relax %
\usepackage{enumitem}  
\setlist[description]{font=\normalfont\itshape}

\usepackage{siunitx}
\DeclareSIUnit\gauss{G}

\usepackage{eurosym}

\usepackage{pgfplots}
\usetikzlibrary{shapes,arrows,fit,calc}
\usepackage[dvipsnames]{xcolor}
\usepackage{pgfplotstable}
\usetikzlibrary{math} %
\usepackage{tikz}
\pgfplotsset{compat=newest} 
\pgfplotsset{plot coordinates/math parser=false} 
\newlength\figureheight 
\newlength\figurewidth

\usepackage{array}
\newcolumntype{L}[1]{>{\raggedright\let\newline\\\arraybackslash\hspace{0pt}}m{#1}}
\newcolumntype{C}[1]{>{\centering\let\newline\\\arraybackslash\hspace{0pt}}m{#1}}
\newcolumntype{R}[1]{>{\raggedleft\let\newline\\\arraybackslash\hspace{0pt}}m{#1}}
\newcolumntype{M}{>{\centering\let\newline\\\arraybackslash\hspace{0pt}}X} %
\newcolumntype{s}{>{\hsize=.4\hsize}X}

\makeatletter
\newcommand{\thickhline}{%
    \noalign {\ifnum 0=`}\fi \hrule height 1pt
    \futurelet \reserved@a \@xhline
}
\newcolumntype{"}{@{\hskip\tabcolsep\vrule width 1pt\hskip\tabcolsep}}
\makeatother


\algnewcommand\algorithmicinput{\textbf{Initialization:}}
\algnewcommand\init{\item[\algorithmicinput]}

\algnewcommand\algorithmicalib{\textbf{Calibration:}}
\algnewcommand\calib{\item[\algorithmicalib]}

\algnewcommand\algorithmicevol{\textbf{Monitoring:}}
\algnewcommand\evol{\item[\algorithmicevol]}

\algdef{SE}[SUBALG]{Indent}{EndIndent}{}{\algorithmicend\ }%
\algtext*{Indent}
\algtext*{EndIndent}

\newcommand{\ie}{{i.e.},~}

\newcommand{\ied}{{i.e.}:~}

\newcommand{\fig}{Fig.~}

\newcommand{\tabb}{Tab.~}

\newcommand{\sect}{Sect.~}

\newcommand{\pg}[1]{\paragraph{\textbf{#1}}\mbox{}}

\definecolor{mycolor1}{rgb}{0.00000,0.44700,0.74100}%
\definecolor{mycolor2}{rgb}{0.85000,0.32500,0.09800}%
\definecolor{mycolor3}{rgb}{0.92900,0.69400,0.12500}
\definecolor{mycolor4}{rgb}{0.290,0.624,0.4}

\definecolor{acrobatYellow}{rgb}{1,1,0}
\definecolor{lightgray}{rgb}{0.83, 0.83, 0.83}
\sethlcolor{lightgray}

\newcommand{\orcid}[1]{\href{https://orcid.org/#1}{\includegraphics[scale=0.5]{img/orcid_16x16.png}}}